\documentclass[a4paper,fleqn]{cas-sc}

\usepackage[numbers]{natbib}

\usepackage[ruled,vlined]{algorithm2e}

\usepackage{xcolor}

\def\tsc#1{\csdef{#1}{\textsc{\lowercase{#1}}\xspace}}
\tsc{WGM}
\tsc{QE}

\begin{document}
\let\WriteBookmarks\relax
\def\floatpagepagefraction{1}
\def\textpagefraction{.001}

\shorttitle{Incorporating MAB with Local Search for MaxSAT}

\shortauthors{Zheng and He et al.}  

\title [mode = title]{Incorporating Multi-armed Bandit with Local Search for MaxSAT}




%

\author[1,2]{Jiongzhi Zheng}
\ead{jzzheng@hust.edu.cn}

\credit{Conceptualization of this study, Methodology, Software, Writing and revision}

\address[1]{School of Computer Science and Technology, Huazhong University of Science and Technology, Wuhan 430074, China}
\address[2]{Hopcroft Center on Computing Science, Huazhong University of Science and Technology, Wuhan 430074, China}
\address[3]{MIS, University of Picardie Jules Verne, Amiens 80039, France}
\address[4]{Artificial Intelligence Research Institute (IIIA), CSIC, Bellaterra 08193, Spain}

\author[1,2]{Kun He}[orcid=0000-0001-7627-4604]
\cormark[1]
\ead{brooklet60@hust.edu.cn}
\cortext[cor1]{Corresponding author.}
\credit{Conceptualization of this study, Methodology, Supervision, Writing and revision}

\author[1,2]{Jianrong Zhou}
\credit{Writing and revision}

\author[1,2]{Yan Jin}
\credit{Writing and revision}

\author[3]{Chu-Min Li}
\credit{Writing and revision}

\author[4]{Felip Manyà}
\credit{Writing and revision}

\begin{abstract}
Partial MaxSAT (PMS) and Weighted PMS (WPMS) are two practical generalizations of the MaxSAT problem. In this paper, we propose a local search algorithm for these problems, called BandHS, which applies two multi-armed bandits to guide the search directions when escaping local optima. One bandit is combined with all the soft clauses to help the algorithm select to satisfy appropriate soft clauses, and the other bandit with all the literals in hard clauses to help the algorithm select appropriate literals to satisfy the hard clauses. These two bandits can improve the algorithm's search ability in both feasible and infeasible solution spaces. We further propose an initialization method for (W)PMS that prioritizes both unit and binary clauses when producing the initial solutions. Extensive experiments demonstrate the excellent performance and generalization capability of our proposed methods, that greatly boost the state-of-the-art local search algorithm, SATLike3.0, and the state-of-the-art SAT-based incomplete solver, NuWLS-c.
\end{abstract}

\begin{keywords}
Maximum satisfiability \sep Local search \sep Multi-armed bandit \sep Hybrid decimation
\end{keywords}

\maketitle

\section{Introduction}
\label{Sec-Intro}
As an optimization extension of the famous Boolean Satisfiability (SAT) decision problem, the Maximum Satisfiability (MaxSAT) problem aims at finding a complete assignment of the Boolean variables to satisfy as many clauses as possible in a given propositional formula in Conjunctive Normal Form (CNF)~\cite{li2021maxsat}. Partial MaxSAT (PMS) is a variant of MaxSAT where the clauses are divided into hard and soft. PMS aims at maximizing the number of satisfied soft clauses with the constraint that all the hard clauses must be satisfied. Associating a positive weight to each soft clause in PMS results in Weighted PMS (WPMS), whose goal is to maximize the total weight of satisfied soft clauses with the same constraint of PMS that all the hard clauses must be satisfied. Both PMS and WPMS, denoted as (W)PMS, have many practical applications such as planning~\cite{Bonet2019}, combinatorial testing~\cite{AMOST22}, group testing~\cite{Ciampiconi2020}, timetabling~\cite{Demirovic2017}, etc.

Existing solvers for (W)PMS can be divided into complete and incomplete according to whether their solutions have optimality guarantees. Complete solvers mainly include branch and bound algorithms~\cite{Li2007,Heras2007,Li2010,Cherif2020,Li2021} and 
SAT-based solvers~\cite{Fu2006,Ansotegui2013,Narodytska2014,Ansotegui2017,Berg2019,Nadel2019}, that solve (W)PMS by iteratively calling an SAT solver. SAT-based solvers are known to be one of the most popular and best-performing categories of methods for (W)PMS, especially for industrial instances. They can be easily modified to be incomplete by returning better solutions when found~\cite{Berg2019,Nadel2019}. Branch and bound algorithms usually incorporate effective inference rules and lower bound evaluation methods. Recently, Li et al.~\cite{Li2021} proposed to apply clause learning techniques~\cite{Li2020} to boost branch and bound MaxSAT solvers, yielding competitive performance than the state-of-the-art SAT-based solvers.

Incomplete algorithms mainly focus on local search methods~\cite{Selman1993,Morris1993,Cha1997,Cai2016,Luo2017,Cai2020,Zheng2022}. Although local search solvers are not as good as SAT-based solvers in solving large industrial instances, they exhibit promising performance in solving random and crafted instances, and the combination of local search methods with complete solvers shows great potential in solving both SAT~\cite{Cai2022} and MaxSAT~\cite{Cai2020,Chu2022nuwls}. Local search algorithms usually flip the Boolean value of a selected variable in each step to explore the solution space. Two essential techniques in local search algorithms are the clause weighting scheme and the variable selection strategy. Recent well-performing local search (W)PMS algorithms have proposed many effective clause weighting schemes~\cite{Cai2014,Cai2016,Lei2018,Cai2020,Chu2022nuwls}. However, the variable selection strategies used in these algorithms are not good enough, especially when 
falling into the local optima. Note that a local optimum indicates that flipping any single variable cannot improve the current solution.

When falling into an infeasible local optimum (i.e., there are falsified hard clauses), these algorithms~\cite{Cai2014,Cai2016,Lei2018,Cai2020,Chu2022nuwls} first randomly select a falsified hard clause and then satisfy it by flipping one of its variables. The random strategy for selecting the falsified hard clause is reasonable, since all the hard clauses should be satisfied. However, when falling into a feasible local optimum (i.e., there are no falsified hard clauses), these algorithms still use the random strategy to determine the soft clause to be satisfied in the current step, which may not be a good strategy for the following reasons: 1) different from hard clauses, not all the soft clauses should be satisfied. 2) the high degree of randomness may lead to a small probability for these algorithms to find a good search direction (satisfying a falsified soft clause corresponds to a search direction).

To handle the above issues, we propose to use a multi-armed bandit (MAB) to help the local search algorithm select to satisfy appropriate soft clauses. MAB is a basic model for reinforcement learning~\cite{Slivkins2019,Lattimore2020}. In an MAB reinforcement learning model, the agent needs to select to pull an arm (i.e., perform an action) at each decision step (i.e., state), which leads to some rewards. The agent uses the rewards to evaluate the benefit of pulling each arm and uses the estimated values to decide the arm to be pulled in each step. In summary, the MAB can be used to help a program learn to select an appropriate item from multiple candidates. The proposed MAB is combined with all the soft clauses in the (W)PMS instance, thus is denoted as \textit{soft} MAB. Each arm in the \textit{soft} MAB corresponds to a soft clause. The \textit{soft} MAB is called to replace the random strategy to decide the soft clause to be satisfied in the current step whenever the search falls into a feasible local optimum.

The \textit{soft} MAB can improve the algorithm's search ability in feasible solution space. To achieve a more comprehensive improvement, we propose a \textit{hard} MAB to further improve the algorithm's search ability in infeasible solution space. Although all the hard clauses should be satisfied, satisfying which literal in each hard clause is a key point. Based on this intuition, we propose to combine the \textit{hard} MAB with all the literals in hard clauses in the (W)PMS instance, so as to help the algorithm select appropriate literals to satisfy the hard clauses and find feasible solutions faster. The new local search algorithm proposed in this paper with both the \textit{hard} and \textit{soft} MABs is denoted as BandHS.

Moreover, inspired by the studies for SAT and MaxSAT that prioritize both unit and binary clauses (i.e., clauses with exactly one and two literals, respectively) over other clauses~\cite{Chao1990,Chvatal1992,Li2006}, we propose a novel decimation approach that prefers to satisfy both unit and binary clauses, denoted as hybrid decimation (HyDeci), for generating the initial assignment in BandHS. The decimation method is a category of incomplete approaches that proceeds by assigning the Boolean value of some (usually one) variables sequentially and simplifies the formula accordingly~\cite{Cai2017}. Decimation approaches that focus on unit clauses have been used in MaxSAT~\cite{Cai2017,Cai2020}. However, it is the first time, to our knowledge, that a decimation method concentrating on both unit and binary clauses is used in MaxSAT. The experiments demonstrate that considering both unit and binary clauses is better than only considering unit clauses.

There are some related studies that apply MAB to MaxSAT and SAT. For example, Goffinet and Ramanujan~\cite{Goffinet2016} proposed an algorithm for MaxSAT, based on Monte-Carlo tree search, where a two-armed bandit is associated with each variable (node in the search tree) to decide the branching direction, i.e., assign which Boolean value to the variable. Lassouaoui et al.~\cite{Lassouaoui2019} proposed to use an MAB model to select the low-level heuristics in a hyper-heuristic framework for MaxSAT, where pulling an arm implies selecting a corresponding low-level heuristic. Cherif et al.~\cite{Cherif2021} proposed to use a two-armed bandit to select the branching heuristic that decides the next variable to branch on in an SAT solver. These methods have made some achievements in solving MaxSAT and SAT. Our work proposes two novel MAB models for (W)PMS that significantly improves the state-of-the-art local search (W)PMS methods. To our knowledge, this is the first time that an MAB model is associated with all the soft clauses, and also the first time that an MAB model is associated with all the literals in hard clauses, in a local search (W)PMS solver.

This paper is an extended and improved version of our conference paper~\cite{Zheng2022}, in which we proposed the BandMaxSAT algorithm with the HyDeci initialization algorithm and the soft \textit{MAB} model. In this paper, we further propose the \textit{hard} MAB model in BandHS and make a more comprehensive empirical analysis of our proposed methods. To evaluate the performance of the proposed methods, we compare BandHS with the state-of-the-art local search (W)PMS algorithm SATLike3.0. The experiments show that BandHS significantly outperforms SATLike3.0 on both PMS and WPMS. We also compare BandHS with BandMaxSAT to show the effectiveness of the \textit{hard} MAB. Moreover, we apply our proposed MAB models to improve the local search component of the state-of-the-art SAT-based incomplete (W)PMS solver, NuWLS-c~\cite{Chu2022nuwls}, which won all the four incomplete tracks of MaxSAT Evaluation 2022. The resulting solver NuWLS-c-Band shows better performance than NuWLS, indicating the excellent generalization capability of our proposed methods.


The main contributions of this work are as follows:
\begin{itemize}
\item We propose to combine an MAB model with all the soft clauses to improve local search (W)PMS algorithms, and propose to combine another MAB model with all the literals in hard clauses to further improve the algorithm's performance. We demonstrate the great potential to use MAB in MaxSAT solving.
\item We propose a novel decimation method for (W)PMS, denoted as HyDeci, that prefers to satisfy both unit and binary clauses while constructing the initial solution. HyDeci can provide high-quality initial assignments, and could be applied to improve other local search MaxSAT algorithms.
\item Extensive experiments show the excellent performance and generalization capability of our proposed methods, that significantly improve the state-of-the-art local search and SAT-based incomplete (W)PMS solvers.
\end{itemize}

The rest of this paper is organized as follows. Section \ref{Sec-Pre} introduces some preliminary concepts. Section \ref{Sec-BandHS} introduces the proposed BandHS algorithm, including the HyDeci initialization method and the proposed \textit{soft} and \textit{hard} MAB models. Section \ref{Sec-Exp} evaluates the performance of the proposed methods. Section \ref{Sec-Con} contains the concluding remarks.

\section{Preliminaries}
\label{Sec-Pre}
Given a set of Boolean variables $\{x_1,...,x_n\}$, a literal is either a variable itself $x_i$ or its negation $\lnot x_i$; a clause is a disjunction of literals, i.e., $c_j = l_{j1} \lor ... \lor l_{j n_j}$, where $n_j$ is the number of literals in clause $c_j$. A Conjunctive Normal Form (CNF) formula $\mathcal{F}$ is a conjunction of clauses, i.e., $\mathcal{F} = c_1 \land ... \land c_m$. A complete assignment $A$ represents a mapping that maps each variable to a value of 1 (true) or 0 (false). A literal $x_i$ (resp. $\lnot x_i$) is satisfied if the current assignment maps $x_i$ to 1 (resp. 0). A clause is satisfied by the current assignment if there is at least one satisfied literal in the clause. 

Given a CNF formula $\mathcal{F}$, SAT is a decision problem that aims at determining whether there is an assignment that satisfies all the clauses in $\mathcal{F}$, and MaxSAT aims at finding an assignment that satisfies as many clauses in $\mathcal{F}$ as possible. Given a CNF formula $\mathcal{F}$ whose clauses are divided into hard and soft, PMS is a variant of MaxSAT that aims at finding an assignment that satisfies all the hard clauses meanwhile maximizing the number of satisfied soft clauses in $\mathcal{F}$, and WPMS is a generalization of PMS where each soft clause is associated with a positive weight. The goal of WPMS is to find an assignment that satisfies all the hard clauses meanwhile maximizes the total weight of satisfied soft clauses in $\mathcal{F}$. In the local search algorithms for MaxSAT, the flipping operator for a variable is an operator that changes its Boolean value.


Given a (W)PMS instance $\mathcal{F}$, a complete assignment $A$ is feasible if it satisfies all the hard clauses in $\mathcal{F}$. The cost of $A$, denoted as $cost(A)$, is set to $+\infty$ for convenience if $A$ is infeasible. Otherwise, $cost(A)$ is equal to the number of falsified soft clauses for PMS, and equal to the total weight of falsified soft clauses for WPMS.

In addition, the effective clause weighting technique is widely used in recent well-performing (W)PMS local search algorithms~\cite{Cai2014,Luo2017,Cai2020,Chu2022nuwls}. Algorithms with this technique associate dynamic weights (independent of the original soft clause weights in WPMS instances) to clauses and use the dynamic weights to guide the search direction. BandHS also applies the clause weighting technique, and maintains dynamic weights to both hard and soft clauses with the clause weighting strategy used in SATLike3.0~\cite{Cai2020}.

Given a (W)PMS instance $\mathcal{F}$, the current assignment $A$, and the dynamic clause weights, the commonly used scoring function for a variable $x$, denoted as $score(x)$, is defined as the increment of the total dynamic weight of satisfied clauses caused by flipping $x$ in $A$. Moreover, a local optimum for (W)PMS indicates that there are no variables with positive \textit{score}. A local optimum is feasible if there are no falsified hard clauses, otherwise it is infeasible.

\section{Methodology}
\label{Sec-BandHS}
BandHS consists of the proposed hybrid decimation (HyDeci) initialization process and the local search process. The proposed \textit{hard} and \textit{soft} MABs are trained and used during the local search process to guide the search directions when escaping the local optima. This section first introduces the main framework of the proposed BandHS algorithm, and then introduces its components, including the proposed two MAB models and the HyDeci method.

\subsection{Main framework of BandHS}
\label{Sec-MainProcess}
Before introducing the main process of BandHS, we first introduce some concepts and definitions. In BandHS, each arm in the \textit{hard} MAB corresponds to a literal in hard clauses, and each arm in the \textit{soft} MAB corresponds to a soft clause. We associate an estimated value with each arm in the MAB models to evaluate the benefits that may yield after being pulled, i.e., satisfying its corresponding literal or soft clause. For each arm $l$ (resp. $i$) in the \textit{hard} (resp. \textit{soft}) MAB, we define $V^h_l$ (resp. $V^s_i$) as its estimated value, initialized to be 1, and $t^h_l$ (resp. $t^s_i$) as the number of times it has been pulled. Intuitively, the larger the estimated value of an arm in the \textit{hard} MAB, satisfying its corresponding literal leads to lesser falsified hard clauses. The larger the estimated value of an arm in the \textit{soft} MAB, satisfying its corresponding soft clause leads to better feasible solutions. Moreover, we define $N^h$ and $N^s$ as the number of times the algorithm falls into infeasible and feasible local optima, respectively.

\begin{algorithm}[t]
\caption{BandHS}
\label{alg:BandHS}
\LinesNumbered 
\KwIn{A (W)PMS instance $\mathcal{F}$, cut-off time \textit{cutoff}, BMS parameter $k$, reward delay steps $d$, reward discount factor $\gamma$, number of sampled arms $ArmNum$, exploration bias parameter $\lambda$}
\KwOut{A feasible assignment $A$ of $\mathcal{F}$, or \textit{no feasible assignment found}}
$A :=$ HyDeci($\mathcal{F}$)\;
initialize $A^* := A$, $A' := A$, $N^s := 0$\;
initialize $H := +\infty$, $H' := +\infty$, $N^h := 0$\;
\While{\textit{running time} $<$ \textit{cutoff}}{
\If{$A$ is feasible $\&$ \textit{cost}($A$) $<$ \textit{cost}($A^*$)}{
$A^* := A$\;
}
\eIf{$D := \{x|score(x)>0\} \neq \emptyset$}{
$v :=$ a variable in $D$ picked by BMS($k$)\;
}
{update\_clause\_weights()\;
\eIf{$\exists$ \textit{falsified hard clauses}}
{
$H :=$ the number of falsified hard clauses in $A$\;
$c :=$ a random falsified hard clause\;
\eIf{$A^*$ is infeasible}
{
update\_hard\_estimated\_value($H,H',d,\gamma$)\;
$N^h := N^h + 1$, $H' := H$\;
$l :=$ PickHardArm($c,N^h,\lambda$)\;
$t^h_l := t^h_l + 1$\;
$v :=$ the variable corresponds to literal $l$\;}
{$v :=$ the variable with the highest \textit{score} in $c$\;}
}
{
update\_soft\_estimated\_value($A,A',A^*,d,\gamma$)\;
$N^s := N^s + 1$, $A' := A$\;
$c :=$ PickSoftArm($ArmNum,N^s,\lambda$)\;
$t^s_c := t^s_c + 1$\;
$v :=$ the variable with the highest \textit{score} in $c$\;
}
}
$A := A$ with $v$ flipped\;
}
\lIf{$A^*$ is feasible}{\textbf{return} $A^*$}
\lElse{\textbf{return} \textit{no feasible assignment found}}
\end{algorithm}

The main process of BandHS is shown in Algorithm \ref{alg:BandHS}. BandHS first uses the HyDeci method (see Algorithm \ref{alg:HyDeci}) to generate an initial assignment (line 1), and then repeatedly selects a variable 
to flip 
until the cut-off time is reached (lines 4-28). When local optima are not reached, BandHS selects a variable to be flipped using the Best from Multiple Selections (BMS) strategy~\cite{Cai2015}. BMS chooses $k$ random variables (with replacement) and returns one with the highest \textit{score} (lines 7-8). When falling into a local optimum, BandHS first updates the dynamic clause weights (by the update\_clause\_weight() function in line 10) according to the clause weighting scheme in SATLike3.0~\cite{Cai2020}, and then selects the variable to be flipped in the current step.

If the local optimum is infeasible, BandHS first randomly selects a falsified hard clause $c$ as the clause to be satisfied (line 13). Then, if no feasible solutions have been found (lines 14-19), the \textit{hard} MAB will be called to pick an arm $l$ among all the arms (i.e., literals) in $c$ by function PickHardArm() (see Algorithm \ref{alg:PickHardArm}). The variable to be flipped in the current step is the variable that corresponds to literal $l$. If BandHS finds any feasible solution, it will not call the \textit{hard} MAB to select the variable to be flipped but selects it greedily according to the scoring function (line 21).

If the local optimum is feasible (lines 22-27), BandHS first calls the \textit{soft} MAB model to select the soft clause $c$ to be satisfied in the current step by function PickSoftArm() (see Algorithm \ref{alg:PickSoftArm}), and then selects to flip the variable with the highest \textit{score} in $c$ in the current step (line 27).

Moreover, note that once an arm is picked, the related information, including the estimated values, the number of pulled times, and the number of times to fall into local optima, will be updated (lines 15-18 and 23-26).

In summary, the \textit{hard} and \textit{soft} MABs are trained and used separately in BandHS. Before finding any feasible solution, BandHS focuses on training and using the \textit{hard} MAB to help it find feasible solutions faster. After finding any feasible solution, BandHS focuses on training and using the \textit{soft} MAB to help it find better solutions. We stop training the \textit{hard} MAB after finding a feasible solution because the transformation of searching between feasible and infeasible solution spaces may mislead the reward function and the estimated values of the arms in the \textit{hard} MAB. As a result, the \textit{hard} and \textit{soft} MABs can improve the search ability of BandHS in infeasible and feasible solution spaces, respectively.

\subsection{The \textit{hard} and \textit{soft} MAB models}
From the above subsection, we know when will the two MAB models be called and how they are used, i.e., used to determine the literal or the soft clause to be satisfied. This subsection introduces details of the MAB models, including how to select the arms to be pulled and how to update the estimated values.

\subsubsection{Arm selection strategy}
BandHS adopts the Upper Confidence Bound method~\cite{Hu2019} to trade-off between exploration and exploitation and select the arms to be pulled. Specifically, the upper confidence bound $U^h_l$ on the estimated value $V^h_l$ of arm $l$ in the \textit{hard} MAB is calculated by the following equation:
\begin{equation}
U^h_l = V^h_l + \lambda \cdot \sqrt{\frac{ln(N^h)}{t^h_l + 1}},
\label{eq:UCBh}
\end{equation}
where $\lambda$ is the exploration bias parameter, $N^h$ is the number of times the algorithm falls into infeasible local optima, and $t^h_l$ is the number of times arm $l$ has been pulled in the \textit{hard} MAB.


Similarly, the upper confidence bound $U^s_i$ on the estimated value $V^s_i$ of arm $i$ in the \textit{soft} MAB is calculated by:
\begin{equation}
U^s_i = V^s_i + \lambda \cdot \sqrt{\frac{ln(N^s)}{t^s_i + 1}},
\label{eq:UCBs}
\end{equation}
where $N^s$ is the number of times the algorithm falls into feasible local optima, and $t^s_i$ is the number of times arm $i$ has been pulled in the \textit{soft} MAB.

The procedures of selecting an arm to be pulled in the \textit{hard} and \textit{soft} MABs, i.e., functions PickHardArm() and PickSoftArm(), are shown in Algorithms \ref{alg:PickHardArm} and \ref{alg:PickSoftArm}, respectively. Since the proposed MABs contain a large number of arms (equal to the number of literals in hard clauses or the number of soft clauses), selecting the best among all the arms is inefficient. Therefore, we propose to use a sampling strategy to reduce the selection scope and improve the algorithm's efficiency. When selecting an arm to be pulled in the \textit{hard} MAB, i.e., selecting a falsified literal in hard clauses, the selection of the random falsified hard clause $c$ (where all the literals are falsified) in line 13 of Algorithm \ref{alg:BandHS} can be regarded as a natural and reasonable sampling strategy, since all the hard clauses should be satisfied. PickHardArm() actually returns an arm (i.e., literal) with the highest upper confidence bound in the input clause $c$.

When selecting an arm to be pulled in the \textit{soft} MAB, the PickSoftArm() function first samples $ArmNum$ (20 by default) candidate arms and then selects an arm with the highest upper confidence bound among the candidates. Note that the \textit{soft} MAB aims at selecting a soft clause to be satisfied in the current step. Thus, the arms corresponding to the soft clauses that are satisfied by the current assignment will not be considered as candidates. Similar sampling strategies have been used in MAB problems~\cite{Ou2019} and some combinatorial optimization problems~\cite{Cai2015}. Experimental results also show that the sampling strategy used in BandHS can significantly improve the algorithm's performance.

\begin{algorithm}[t]
\caption{PickHardArm($c,N^h,\lambda$)}
\label{alg:PickHardArm}
\LinesNumbered 
\KwIn{A random falsified hard clause $c$, number of times to fall into an infeasible local optimum $N^h$, exploration bias parameter $\lambda$}
\KwOut{The arm selected to be pulled $l$}
initialize $U^{h}_* := -\infty$\;
\For{each literal $j$ in $c$}{
calculate $U^h_j$ according to Eq. \ref{eq:UCBh}\;
\lIf{$U^h_j > U^h_*$}{$U^{h}_* := U^h_j$, $l := j$}
}
\textbf{return} $l$\;
\end{algorithm}

\begin{algorithm}[t]
\caption{PickSoftArm($ArmNum,N^s,\lambda$)}
\label{alg:PickSoftArm}
\LinesNumbered 
\KwIn{Number of sampled arms $ArmNum$, number of times to fall into a feasible local optimum $N^s$, exploration bias parameter $\lambda$}
\KwOut{The arm selected to be pulled $c$}
initialize $U^{s}_* := -\infty$\;
\For{$i := 1$ to $ArmNum$}{
$j :=$ a random falsified soft clause\;
calculate $U^s_j$ according to Eq. \ref{eq:UCBs}\;
\lIf{$U^s_j > U^s_*$}{$U^{s}_* := U^s_j$, $c := j$}
}
\textbf{return} $c$\;
\end{algorithm}

\subsubsection{Estimated value updating strategy}
In general, pulling a high-quality arm in the \textit{hard} MAB leads to lesser falsified hard clauses and can help the algorithm find feasible solutions faster, and pulling a high-quality arm in the \textit{soft} MAB leads to better feasible solutions. Therefore, we use the changing of the number of falsified hard clauses and the changing of the cost values (see Section \ref{Sec-Pre}) to respectively design the reward functions for updating the estimated values of the arms in the \textit{hard} and \textit{soft} MABs.

Specifically, suppose $H'$ and $H$ are the numbers of falsified hard clauses of the 
previous and current infeasible local optimal solutions, respectively, and $l$ is the 
previous pulled arm in the \textit{hard} MAB, a simple reward for pulling $l$ can be set to $H' - H$. However, reducing the number of falsified hard clauses from 20 to 10 is much harder and more meaningful than reducing it from 1000 to 990. Thus, the rewards of these two cases should not be the same. To address this issue, we define the reward as follows:
\begin{equation}
r^h(H,H') = (H'-H)/H'.
\label{eq:rh}
\end{equation}

From Eq. \ref{eq:rh}, we can observe that suppose $H'-H$ is constant, the smaller the value of $H'$, the more rewards the action of pulling the 
previous arm in the \textit{hard} MAB can yield, which is reasonable and intuitive.

Similarly, suppose $A'$ and $A$ are the 
previous and current feasible local optimal solutions, respectively, and $c$ is the 
previous pulled arm in the \textit{soft} MAB, the reward of pulling arm $c$ is defined as follows:
\begin{equation}
r^s(A,A',A^*) = \frac{cost(A') - cost(A)}{cost(A') - cost(A^*) + 1},
\label{eq:rs}
\end{equation}
where $A^*$ is the best solution found so far, i.e., suppose $cost(A') - cost(A)$ is constant, the closer $cost(A')$ and $cost(A^*)$, the more rewards the action of pulling the 
previous arm in the \textit{soft} MAB can yield.

Moreover, since the arms in the \textit{hard} MAB are connected by the hard clauses, and the arms in the \textit{soft} MAB are connected by the variables, we assume that the arms in our MAB models are not independent of each other. We also believe that the improvement (or deterioration) of $H$ over $H'$ or $A$ over $A'$ may not only be due to the 
previous action, but also to earlier actions. Hence, we apply the delayed reward method~\cite{Arya2019} to update the estimated value of the latest $d$ (20 by default) pulled arms once a reward is obtained.

Specifically, suppose $H'$ and $H$ are the numbers of falsified hard clauses of the previous and current infeasible local optimal solutions, and $\{l_1,...,l_d\}$ is the set of the latest $d$ pulled arms ($l_d$ is the most recent one) in the \textit{hard} MAB. Then, the estimated values of the $d$ arms in the \textit{hard} MAB are updated as follows:
\begin{equation}
V^h_{l_i} = V^h_{l_i} + \gamma^{d-i} \cdot r^h(H,H'), ~~i \in \{1,...,d\},
\label{eq:updateVh}
\end{equation}
where $\gamma$ is the reward discount factor and $r^h(H,H')$ is calculated by Eq.~\ref{eq:rh}.

Similarly, suppose $A'$ and $A$ are the previous and current feasible local optimal solutions, respectively, $A^*$ is the best solution found so far, and $\{a_1,...,a_d\}$ is the set of the latest $d$ pulled arms ($a_d$ is the most recent one) in the \textit{soft} MAB. The estimated values of the $d$ arms in the \textit{soft} MAB are updated as follows:
\begin{equation}
V^s_{a_i} = V^s_{a_i} + \gamma^{d-i} \cdot r^s(A,A',A^*), ~~i \in \{1,...,d\},
\label{eq:updateVs}
\end{equation}
where $r^s(A,A',A^*)$ is calculated by Eq.~\ref{eq:rs}.

The functions update\_hard\_estimated\_values() and update\_soft\_estimated\_values() in lines 15 and 23 of Algorithm \ref{alg:BandHS} actually update the estimated values of the latest pulled $d$ arms according to Eqs. \ref{eq:updateVh} and \ref{eq:updateVh}, respectively.


\subsection{Hybrid decimation}
Finally, we introduce the proposed HyDeci initialization method, an effective decimation method that prefers to satisfy both unit and binary clauses. Since the clauses with shorter lengths are easier to be falsified, preferring to satisfy shorter clauses can reduce the number of falsified clauses, which results in high-quality initial assignments. The procedure of HyDeci is shown in Algorithm \ref{alg:HyDeci}. We use SIMPLIFY to refer to the process of simplifying the formula after assigning a value to a variable.

\begin{algorithm}[t]
\caption{HyDeci($\mathcal{F}$)}
\label{alg:HyDeci}
\LinesNumbered 
\KwIn{A (W)PMS instance $\mathcal{F}$}
\KwOut{A complete assignment $A$ of variables in $\mathcal{F}$}
\While{$\exists$ unassigned variables}{
\If{$\exists$ hard unit clauses}{
$c :=$ a random hard unit clause, satisfy $c$ and SIMPLIFY\;
}
\ElseIf{$\exists$ soft unit clauses}{
$c :=$ a random soft unit clause, satisfy $c$ and SIMPLIFY\;
}
\ElseIf{$\exists$ hard binary clauses}{
$c :=$ a random hard binary clause\;
$l :=$ a greedily selected unassigned literal in $c$, satisfy $l$ and SIMPLIFY\;
}
\ElseIf{$\exists$ soft binary clauses}{
$c :=$ a random soft binary clause\;
$l :=$ a greedily selected unassigned literal in $c$, satisfy $l$ and SIMPLIFY\;
}
\Else{$v :=$ a random unassigned variable, assign $v$ a random value and SIMPLIFY\;
}
}
\textbf{return} the resulting complete assignment $A$\;
\end{algorithm}

HyDeci generates the initial complete assignment iteratively. In each iteration, HyDeci assigns the value of exactly one variable. When there are unit clauses, HyDeci samples a random unit clause (hard clauses take precedence) and then satisfies it. 
When there is no unit clause but are binary ones, HyDeci first samples a random binary clause $c$ (hard clauses take precedence), and then selects one of the two unassigned literals in $c$ and satisfies it according to a greedy strategy, that is, preferring to satisfy the literal whose satisfaction leads to more satisfied soft clauses (or to a larger total weight of satisfied soft clauses). When there are no unit and binary clauses, HyDeci randomly selects an unassigned variable and randomly assigns a Boolean value to it.

In summary, the main improvement of the proposed HyDeci algorithm over the 
existing decimation approaches~\cite{Cai2017,Cai2020} is that HyDeci not only concentrates on unit clauses but also on binary clauses.

\section{Experiments}
\label{Sec-Exp}
In this section, we first compare BandHS with the state-of-the-art local search (W)PMS algorithm, SATLike3.0~\cite{Cai2020}. Then, we apply the proposed MAB models to improve the local search component of the state-of-the-art SAT-based incomplete (W)PMS solver, NuWLS-c~\cite{Chu2022nuwls}, winner of all the four incomplete tracks of MaxSAT Evaluation (MSE) 2022, and compare the resulting solver NuWLS-c-Band with some effective SAT-based incomplete solvers, including NuWLS-c, SATLike-c~\cite{Lei2021}, Loandra~\cite{Berg2019}, and TT-Open-WBO-Inc~\cite{Nadel2019}. Moreover, we do thorough ablation studies to evaluate the effectiveness of components and strategies in BandHS, including the HyDeci initialization method, the \textit{hard} and \textit{soft} MAB models, the sampling strategy for selecting an arm to be pulled, and the delayed reward method.

\subsection{Experimental setup}
BandHS is implemented in C++ and compiled by g++. Our experiments were performed on a server using an Intel® Xeon® E5-2650 v3 2.30 GHz 10-core CPU and 256 GB RAM, running Ubuntu 16.04 Linux operation system. We tested the algorithms on all the (W)PMS instances from the incomplete track of the 
five recent MaxSAT Evaluations (MSE), i.e., MSE2018 to MSE2022. Note that we denote the benchmarks that contain all the PMS/WPMS instances from the incomplete track of MSE2022 as PMS\_2022/WPMS\_2022, and so on. Each instance is calculated once by each algorithm with a time limit of 300 seconds, which is consistent with the settings in MSEs. The best results in the tables appear in bold.

The parameters in BandHS include the BMS parameter $k$, the reward delay steps $d$, the reward discount factor $\gamma$, the number of sampled arms in the \textit{soft} MAB $ArmNum$, and the exploration bias parameter $\lambda$. The default settings of these parameters are as follows: $k = 15$, $d = 20$, $\gamma = 0.9$, $ArmNum = 20$, $\lambda = 1$. The code of BandHS is available at https://github.com/JHL-HUST/BandHS/.

\begin{table}[t]
\footnotesize
\centering
\caption{Comparison of BandHS and SATLike3.0.}
\label{table-BandHS-SL}
\begin{tabular}{lrrrrrrr} \bottomrule
\multirow{2}{*}{Benchmark} & \multirow{2}{*}{\#inst.} &  & \multicolumn{2}{c}{BandHS} &  & \multicolumn{2}{c}{SATLike3.0} \\ \cline{4-5} \cline{7-8} 
                           &                          &  & \#win.          & time     &  & \#win.         & time          \\ \hline
PMS\_2018                  & 153                      &  & \textbf{111}    & 75.65    &  & 57             & 81.70         \\
PMS\_2019                  & 299                      &  & \textbf{205}    & 65.64    &  & 138            & 51.39         \\
PMS\_2020                  & 262                      &  & \textbf{179}    & 63.95    &  & 116            & 58.37         \\
PMS\_2021                  & 155                      &  & \textbf{111}    & 62.08    &  & 58             & 46.94         \\
PMS\_2022                  & 179                      &  & \textbf{132}    & 75.54    &  & 60             & 68.12         \\
WPMS\_2018                 & 172                      &  & \textbf{116}    & 107.41   &  & 52             & 72.10         \\
WPMS\_2019                 & 297                      &  & \textbf{209}    & 102.38   &  & 104            & 78.61         \\
WPMS\_2020                 & 253                      &  & \textbf{176}    & 111.49   &  & 83             & 76.56         \\
WPMS\_2021                 & 151                      &  & \textbf{90}     & 134.58   &  & 54             & 92.74         \\
WPMS\_2022                 & 197                      &  & \textbf{121}    & 133.07   &  & 61             & 107.18    \\ \toprule   
\end{tabular}
\end{table}
\begin{table}[t]
\footnotesize
\centering
\caption{Comparison of BandHS and SATLike3.0 on each PMS instance class.}
\label{table_PMS_class}
\begin{tabular}{lrrrrrrr} \bottomrule
\multirow{2}{*}{Instance class}         & \multirow{2}{*}{\#inst.} &  & \multicolumn{2}{c}{BandHS}     &  & \multicolumn{2}{c}{SATLike3.0} \\ \cline{4-5} \cline{7-8} 
                                        &                          &  & \#win.       & time            &  & \#win.        & time           \\ \hline
aes                                     & 6                        &  & \textbf{6}   & 84.09           &  & 2             & 136.81         \\
atcoss                                  & 14                       &  & \textbf{1}   & 39.40           &  & 0             & 0.00           \\
bcp                                     & 24                       &  & \textbf{20}  & 104.14          &  & 11            & 126.47         \\
causal-discovery                        & 3                        &  & 3            & \textbf{3.83}   &  & 3             & 5.40           \\
close\_solutions                        & 14                       &  & 6            & 46.76           &  & \textbf{9}    & 35.75          \\
decision-tree                           & 31                       &  & \textbf{30}  & 31.96           &  & 6             & 115.48         \\
des                                     & 13                       &  & 0            & 0.00            &  & \textbf{2}    & 89.65          \\
extension-enforcement                   & 19                       &  & \textbf{15}  & 128.35          &  & 14            & 86.49          \\
fault-diagnosis                         & 8                        &  & \textbf{8}   & 20.46           &  & 0             & 0.00           \\
gen-hyper-tw                            & 37                       &  & 24           & \textbf{77.92}  &  & 24            & 115.18         \\
hs-timetabling                          & 1                        &  & \textbf{1}   & 2.74            &  & 0             & 0.00           \\
large-graph-commmunity                  & 3                        &  & \textbf{3}   & 7.00            &  & 2             & 10.52          \\
logic-synthesis                         & 1                        &  & \textbf{1}   & 117.78          &  & 0             & 0.00           \\
maxclique \& maxcut                     & 68                       &  & 63           & 28.57           &  & \textbf{67}   & 2.62           \\
MaximumCommonSub-GraphExtraction        & 25                       &  & 17           & 39.39           &  & \textbf{22}   & 41.19          \\
MaxSATQueriesinInterpretableClassifiers & 42                       &  & 26           & 58.14           &  & \textbf{28}   & 72.02          \\
mbd                                     & 6                        &  & 3            & \textbf{173.16} &  & 3             & 241.39         \\
min-fill                                & 17                       &  & \textbf{12}  & 44.21           &  & 6             & 118.00         \\
optic                                   & 17                       &  & \textbf{16}  & 112.41          &  & 1             & 154.25         \\
phylogenetic-trees                      & 14                       &  & \textbf{4}   & 158.88          &  & 0             & 0.00           \\
pseudoBoolean                           & 11                       &  & \textbf{1}   & 295.01          &  & 0             & 0.00           \\
railroad\_reisch                        & 9                        &  & \textbf{9}   & 72.53           &  & 5             & 6.57           \\
railway-transport                       & 4                        &  & \textbf{2}   & 73.20           &  & 1             & 298.83         \\
ramsey                                  & 14                       &  & 14           & 2.89            &  & 14            & \textbf{0.11}  \\
ran-scp                                 & 14                       &  & \textbf{14}  & 89.21           &  & 1             & 262.36         \\
scheduling                              & 5                        &  & \textbf{3}   & 128.80          &  & 2             & 128.64         \\
scheduling\_xiaojuan                    & 20                       &  & \textbf{20}  & 116.38          &  & 1             & 179.28         \\
SeanSafarpour                           & 13                       &  & \textbf{10}  & 111.89          &  & 6             & 100.73         \\
set-covering                            & 9                        &  & \textbf{8}   & 98.59           &  & 1             & 122.93         \\
setcover-rail\_zhendong                 & 4                        &  & 2            & 1.04            &  & \textbf{4}    & 84.80          \\
treewidth-computation                   & 9                        &  & \textbf{8}   & 102.24          &  & 6             & 70.96          \\
uaq                                     & 20                       &  & \textbf{20}  & 50.34           &  & 18            & 46.98          \\
uaq\_gazzarata                          & 4                        &  & \textbf{4}   & 130.02          &  & 1             & 108.16         \\
xai-mindset2                            & 19                       &  & \textbf{15}  & 52.17           &  & 0             & 0.00           \\ \hline
Total                                   & 518                      &  & \textbf{389} & 72.08           &  & 260           & 58.32     \\ \toprule    
\end{tabular}
\end{table}
\begin{table}[t]
\footnotesize
\centering
\caption{Comparison of BandHS and SATLike3.0 on each WPMS instance class.}
\label{table_WPMS_class}
\begin{tabular}{lrrrrrrr} \bottomrule
\multirow{2}{*}{Instance class}         & \multirow{2}{*}{\#inst.} &  & \multicolumn{2}{c}{BandMaxSAT} &  & \multicolumn{2}{c}{SATLike3.0} \\ \cline{4-5} \cline{7-8} 
                                        &                          &  & \#win.        & time           &  & \#win.       & time            \\ \hline
abstraction-refinement                  & 10                       &  & \textbf{10}   & 155.11         &  & 0            & 0.00            \\
af-synthesis                            & 33                       &  & \textbf{33}   & 141.17         &  & 0            & 0.00            \\
BTBNSL-Rounded                          & 26                       &  & \textbf{20}   & 56.72          &  & 6            & 39.73           \\
causal-discovery                        & 24                       &  & \textbf{20}   & 53.75          &  & 12           & 12.59           \\
cluster-expansion                       & 20                       &  & \textbf{11}   & 67.20          &  & 9            & 68.97           \\
correlation-clustering                  & 46                       &  & \textbf{27}   & 154.62         &  & 19           & 134.85          \\
decision-tree                           & 36                       &  & \textbf{22}   & 150.04         &  & 21           & 136.25          \\
hs-timetabling                          & 13                       &  & \textbf{3}    & 122.41         &  & 1            & 248.98          \\
lisbon-wedding                          & 21                       &  & \textbf{10}   & 131.19         &  & 6            & 170.70          \\
maxcut                                  & 29                       &  & \textbf{29}   & 6.38           &  & 27           & 1.62            \\
max-realizability                       & 13                       &  & \textbf{10}   & 19.84          &  & 8            & 65.04           \\
MaxSATQueriesinInterpretableClassifiers & 37                       &  & \textbf{20}   & 96.83          &  & 18           & 100.87          \\
metro                                   & 2                        &  & 1             & \textbf{87.46} &  & 1            & 183.34          \\
MinimumWeightDominatingSetProblem       & 7                        &  & 6             & \textbf{87.67} &  & 6            & 105.84          \\
min-width                               & 46                       &  & \textbf{40}   & 166.79         &  & 6            & 186.30          \\
mpe                                     & 22                       &  & \textbf{22}   & 114.89         &  & 6            & 93.08           \\
railroad\_reisch                        & 6                        &  & \textbf{6}    & 96.77          &  & 1            & 17.14           \\
railway-transport                       & 4                        &  & \textbf{2}    & 161.96         &  & 1            & 107.09          \\
ramsey                                  & 12                       &  & 7             & 59.81          &  & \textbf{12}  & 32.44           \\
RBAC                                    & 79                       &  & \textbf{56}   & 146.84         &  & 24           & 129.35          \\
relational-inference                    & 2                        &  & 1             & \textbf{77.90} &  & 1            & 212.09          \\
rgg                                     & 1                        &  & 1             & 219.11         &  & 1            & \textbf{216.47} \\
scSequencing\_Mehrabadi                 & 14                       &  & 4             & 216.50         &  & \textbf{10}  & 53.07           \\
set-covering                            & 13                       &  & \textbf{13}   & 69.90          &  & 1            & 42.34           \\
spot5                                   & 7                        &  & \textbf{6}    & 100.58         &  & 1            & 36.09           \\
staff-scheduling                        & 11                       &  & \textbf{10}   & 149.93         &  & 1            & 150.77          \\
tcp                                     & 13                       &  & 5             & 55.79          &  & \textbf{11}  & 104.62          \\
timetabling                             & 19                       &  & \textbf{11}   & 199.74         &  & 4            & 138.80          \\ \hline
Total                                   & 566                      &  & \textbf{406}  & 114.88         &  & 214          & 89.22    \\ \toprule      
\end{tabular}
\end{table}
\begin{table}[t]
\footnotesize
\centering
\caption{Comparison of NuWLS-c-Band and some state-of-the-art SAT-based incomplete solvers, including NuWLS-c, SATLike-c, Loandra, and TT-Open-WBO-Inc (TT-OWI). The results are expressed by the scoring function used in MSEs.}
\label{table-BandHS-c}
\begin{tabular}{lrrrrr} \bottomrule
Benchmark  & NuWLS-c-Band    & NuWLS-c         & SATLike-c & Loandra & TT-OWI \\ \hline
WPMS\_2018 & \textbf{0.9187} & 0.9102          & 0.8868    & 0.8771  & 0.8975 \\
WPMS\_2019 & 0.9149          & \textbf{0.9180} & 0.8804    & 0.8456  & 0.9003 \\
WPMS\_2020 & \textbf{0.8966} & 0.8939          & 0.8589    & 0.8000  & 0.8617 \\
WPMS\_2021 & \textbf{0.8378} & 0.8260          & 0.7765    & 0.7783  & 0.7686 \\
WPMS\_2022 & \textbf{0.8406} & 0.8372          & 0.7861    & 0.8136  & 0.7756 \\
PMS\_2018  & \textbf{0.8745} & 0.8649          & 0.8331    & 0.7725  & 0.8307 \\
PMS\_2019  & 0.8921          & \textbf{0.8976} & 0.8675    & 0.7768  & 0.8646 \\
PMS\_2020  & \textbf{0.8886} & 0.8825          & 0.8414    & 0.8135  & 0.8492 \\
PMS\_2021  & \textbf{0.8914} & 0.8906          & 0.8457    & 0.7646  & 0.8364 \\
PMS\_2022  & \textbf{0.8926} & 0.8786          & 0.8562    & 0.7536  & 0.8613 \\ \toprule
\end{tabular}
\end{table}

\subsection{Comparison of BandHS and SATLike3.0}
We first compare BandHS with the state-of-the-art (W)PMS local search algorithm, SATLike3.0~\cite{Cai2020}, on all the tested instances. The results are shown in Table \ref{table-BandHS-SL}. Column \textit{\#inst.} indicates the number of instances in each benchmark. Column \textit{\#win.} indicates the number of instances in which the algorithm yields the best solution among all the algorithms in the table. Column \textit{time} represents the average running time (in seconds) to yield the \textit{\#win.} instances.

As shown by the results in Table \ref{table-BandHS-SL}, BandHS significantly outperforms SATLike3.0 for both PMS and WPMS. Specifically, the number of \textit{\#win.} instances of BandHS is 49-120\% greater than that of SATLike3.0 for PMS, and 67-123\% greater than that of SATLike3.0 for WPMS, indicating a significant improvement.

To obtain a more detailed comparison between BandHS and SATLike3.0 and evaluate the performance of BandHS on different instance classes, we collect all the tested instances (duplicated instances are removed) and compare BandHS and SATLike3.0 on each instance class. Ties of these two algorithms with the same number of \textit{\#win.} instances are broken by selecting the one with less running time (as the rules in MSEs). The results on the PMS and WPMS instance classes are shown in Tables \ref{table_PMS_class} and \ref{table_WPMS_class}, respectively. Note that we remove the instance classes that both BandHS and SATLike3.0 can not yield feasible solutions. The results show that BandHS outperforms SATLike3.0 on most classes of both PMS and WPMS instances. Specifically, for all the 34 (resp. 28) classes of PMS (resp. WPMS) instances, BandHS outperforms SATLike3.0 on 27 (resp. 24) classes, indicating the excellent robustness of BandHS. Moreover, BandHS shows excellent performance in solving some instance classes, such as \textit{fault-diagnosis}, \textit{optic}, \textit{ran-scp}, \textit{scheduling\_xiaojuan}, and \textit{set-covering} in Table \ref{table_PMS_class}, and \textit{abstraction-refinement}, \textit{af-synthesis}, \textit{min-width}, \textit{set-covering}, and \textit{staff-scheduling} in Table \ref{table_WPMS_class}.

\subsection{Comparison with SAT-based solvers}
We then apply the proposed MAB models to improve the local search component of NuWLS-c~\cite{Chu2022nuwls} and denote the resulting solver as NuWLS-c-Band. Actually, NuWLS-c-Band replaces the local search algorithm in NuWLS-c with the proposed BandHS algorithm. We compare NuWLS-c-Band with some effective SAT-based incomplete solvers that showed excellent performance in recent MSEs, including NuWLS-c, SATLike-c, Loandra, and TT-Open-WBO-Inc (TT-OWI). We apply the scoring function used in MSEs to evaluate the performance of these five solvers. The scoring function actually indicates how close the solutions are to the best-known solutions. Specifically, suppose $C_{BKS}$ is the cost of the best-known solution of an instance recorded in MSEs, $C_i$ is the cost of the solution found by the $i$-th solver ($i \in \{1,2,3,4,5\}$) in our experiments, the score of solver $i$ for this instance is $\frac{\mathop{min}(C_{BKS},C_1,C_2,C_3,C_4,C_5) + 1}{C_i + 1} \in [0,1]$ (resp. 0) if the solution found by solver $i$ is feasible (resp. infeasible). Finally, the score of a solver for a benchmark is the average value of the scores for all the instances in the benchmark.

The comparison results of these five solvers are shown in Table \ref{table-BandHS-c}. We can observe that NuWLS-c-Band yields the highest score on eight among all the ten benchmarks, demonstrating the excellent performance and generalization capability of our proposed MAB methods.

\begin{table}[t]
\footnotesize
\centering
\caption{Comparison of BandHS and BandMaxSAT.}
\label{table-BandHS-BandMS}
\begin{tabular}{lrrrrrrr} \bottomrule
\multirow{2}{*}{Benchmark} & \multirow{2}{*}{\#inst.} &  & \multicolumn{2}{c}{BandHS} &  & \multicolumn{2}{c}{BandMaxSAT} \\ \cline{4-5} \cline{7-8} 
                           &                          &  & \#win.          & time      &  & \#win.         & time          \\ \hline
WPMS\_2018                 & 172                      &  & \textbf{113}    & 95.33     &  & 104            & 100.62        \\
WPMS\_2019                 & 297                      &  & \textbf{200}    & 98.49     &  & 183            & 100.82        \\
WPMS\_2020                 & 253                      &  & \textbf{171}    & 106.55    &  & 158            & 109.24        \\
WPMS\_2021                 & 151                      &  & \textbf{95}     & 126.15    &  & 78             & 136.28        \\
WPMS\_2022                 & 197                      &  & \textbf{116}    & 123.65    &  & 104            & 124.95        \\
PMS\_2018                  & 153                      &  & \textbf{102}    & 78.36     &  & 93             & 83.37         \\
PMS\_2019                  & 299                      &  & \textbf{200}    & 68.41     &  & 189            & 73.33         \\
PMS\_2020                  & 262                      &  & \textbf{179}    & 62.79     &  & 165            & 63.11         \\
PMS\_2021                  & 155                      &  & \textbf{101}    & 60.52     &  & 98             & 60.87         \\
PMS\_2022                  & 179                      &  & \textbf{120}    & 75.06     &  & 116            & 75.45        \\ \toprule
\end{tabular}
\end{table}
\begin{table}[t]
\footnotesize
\centering
\caption{Comparison of BandHS$_{fast}$ and BandMaxSAT$_{fast}$, two variants of BandHS and BandMaxSAT.}
\label{table-BandHS-fast}
\begin{tabular}{lrrrrrrr} \bottomrule
\multirow{2}{*}{Benchmark} & \multirow{2}{*}{\#inst.} &  & \multicolumn{2}{c}{BandHS$_{fast}$} &  & \multicolumn{2}{c}{BandMaxSAT$_{fast}$} \\ \cline{4-5} \cline{7-8} 
                           &                          &  & \#win.              & time$_1$       &  & \#win.               & time$_1$         \\ \hline
WPMS\_2018                 & 172                      &  & \textbf{128}        & 43.37          &  & 116                  & 45.79            \\
WPMS\_2019                 & 297                      &  & \textbf{223}        & 44.37          &  & 217                  & 46.43            \\
WPMS\_2020                 & 253                      &  & \textbf{190}        & 38.33          &  & 182                  & 41.80            \\
WPMS\_2021                 & 151                      &  & 93         & 53.72          &  & 93                   & \textbf{52.74}            \\
WPMS\_2022                 & 197                      &  & \textbf{129}        & 44.90          &  & 118                  & 45.57            \\
PMS\_2018                  & 153                      &  & 96                  & 54.47          &  & \textbf{97}          & 58.07            \\
PMS\_2019                  & 299                      &  & \textbf{202}        & 61.80          &  & 197                  & 62.14            \\
PMS\_2020                  & 262                      &  & \textbf{181}        & 56.60          &  & 176                  & 58.58            \\
PMS\_2021                  & 155                      &  & 94                  & 49.90          &  & \textbf{97}          & 51.42            \\
PMS\_2022                  & 179                      &  & \textbf{121}        & 40.64          &  & 111                  & 42.47     \\ \toprule      
\end{tabular}
\end{table}
\begin{table}[t]
\footnotesize
\centering
\caption{Comparison of BandHS and its variant BandHS$_{\text{no-soft}}$.}
\label{table-BandHS-soft}
\begin{tabular}{lrrrrrrr} \bottomrule
\multirow{2}{*}{Benchmark} & \multirow{2}{*}{\#inst.} &  & \multicolumn{2}{c}{BandHS} &  & \multicolumn{2}{c}{BandHS$_{\text{no-soft}}$} \\ \cline{4-5} \cline{7-8} 
                           &                          &  & \#win.          & time     &  & \#win.             & time              \\ \hline
PMS\_2018                  & 153                      &  & \textbf{108}    & 67.53    &  & 88                 & 57.81             \\
PMS\_2019                  & 299                      &  & \textbf{207}    & 63.77    &  & 173                & 49.46             \\
PMS\_2020                  & 262                      &  & \textbf{179}    & 61.47    &  & 148                & 50.62             \\
PMS\_2021                  & 155                      &  & \textbf{112}    & 62.87    &  & 83                 & 29.10             \\
PMS\_2022                  & 179                      &  & \textbf{128}    & 72.19    &  & 87                 & 47.42             \\
WPMS\_2018                 & 172                      &  & \textbf{116}    & 106.31   &  & 72                 & 69.63             \\
WPMS\_2019                 & 297                      &  & \textbf{209}    & 101.27   &  & 131                & 82.37             \\
WPMS\_2020                 & 253                      &  & \textbf{180}    & 110.02   &  & 110                & 77.73             \\
WPMS\_2021                 & 151                      &  & \textbf{95}     & 134.05   &  & 56                 & 92.90             \\
WPMS\_2022                 & 197                      &  & \textbf{122}    & 125.43   &  & 72                 & 102.81        \\ \toprule   
\end{tabular}
\end{table}
\begin{table}[t]
\footnotesize
\centering
\caption{Comparison of BandHS and its variant BandHS$_{\text{no-binary}}$.}
\label{table-BandHS-binary}
\begin{tabular}{lrrrrrrr} \bottomrule
\multirow{2}{*}{Benchmark} & \multirow{2}{*}{\#inst.} &  & \multicolumn{2}{c}{BandHS} &  & \multicolumn{2}{c}{BandHS$_{\text{no-binary}}$} \\ \cline{4-5} \cline{7-8} 
                           &                          &  & \#win.          & time     &  & \#win.                 & time            \\ \hline
PMS\_2018                  & 153                      &  & \textbf{106}    & 79.96    &  & 105                    & 83.50           \\
PMS\_2019                  & 299                      &  & \textbf{204}    & 70.00    &  & 203                    & 70.51           \\
PMS\_2020                  & 262                      &  & \textbf{182}    & 69.71    &  & 180                    & 70.04           \\
PMS\_2021                  & 155                      &  & \textbf{99}     & 67.13    &  & 98                     & 41.42           \\
PMS\_2022                  & 179                      &  & 118             & 80.59    &  & \textbf{124}           & 77.17           \\
WPMS\_2018                 & 172                      &  & \textbf{104}    & 111.56   &  & 92                     & 86.91           \\
WPMS\_2019                 & 297                      &  & \textbf{201}    & 105.48   &  & 162                    & 93.61           \\
WPMS\_2020                 & 253                      &  & \textbf{171}    & 114.94   &  & 139                    & 96.26           \\
WPMS\_2021                 & 151                      &  & \textbf{97}     & 137.58   &  & 72                     & 110.50          \\
WPMS\_2022                 & 197                      &  & \textbf{119}    & 121.71   &  & 104                     & 113.99       \\ \toprule  
\end{tabular}
\end{table}
\begin{table}[t]
\footnotesize
\centering
\caption{Comparison of BandHS and its variant BandHS$_{\text{no-sample}}$.}
\label{table-BandHS-sample}
\begin{tabular}{lrrrrrrr} \bottomrule
\multirow{2}{*}{Benchmark} & \multirow{2}{*}{\#inst.} &  & \multicolumn{2}{c}{BandHS} &  & \multicolumn{2}{c}{BandHS$_{\text{no-sample}}$} \\ \cline{4-5} \cline{7-8} 
                           &                          &  & \#win.          & time     &  & \#win.              & time               \\ \hline
PMS\_2018                  & 153                      &  & \textbf{105}    & 72.98    &  & 91                  & 59.35              \\
PMS\_2019                  & 299                      &  & \textbf{208}    & 63.81    &  & 169                 & 58.19              \\
PMS\_2020                  & 262                      &  & \textbf{179}    & 65.41    &  & 148                 & 56.98              \\
PMS\_2021                  & 155                      &  & \textbf{114}    & 59.63    &  & 90                  & 50.72              \\
PMS\_2022                  & 179                      &  & \textbf{123}    & 71.38    &  & 103                 & 62.11              \\
WPMS\_2018                 & 172                      &  & \textbf{112}    & 96.32    &  & 77                  & 83.81              \\
WPMS\_2019                 & 297                      &  & \textbf{196}    & 91.11    &  & 157                 & 88.31              \\
WPMS\_2020                 & 253                      &  & \textbf{169}    & 105.57   &  & 127                 & 98.00              \\
WPMS\_2021                 & 151                      &  & \textbf{77}     & 128.64   &  & 70                  & 127.08             \\
WPMS\_2022                 & 197                      &  & \textbf{116}    & 120.73   &  & 82                  & 113.71          \\ \toprule  
\end{tabular}
\end{table}
\begin{table}[t]
\footnotesize
\centering
\caption{Comparison of BandHS and its variant BandHS$_{\text{no-delay}}$.}
\label{table-BandHS-delay}
\begin{tabular}{lrrrrrrr} \bottomrule
\multirow{2}{*}{Benchmark} & \multirow{2}{*}{\#inst.} &  & \multicolumn{2}{c}{BandHS} &  & \multicolumn{2}{c}{BandHS$_{\text{no-delay}}$} \\ \cline{4-5} \cline{7-8} 
                           &                          &  & \#win.          & time     &  & \#win.             & time               \\ \hline
PMS\_2018                  & 153                      &  & \textbf{91}     & 76.37    &  & 89                 & 78.71              \\
PMS\_2019                  & 299                      &  & \textbf{186}    & 66.73    &  & 176                & 65.51              \\
PMS\_2020                  & 262                      &  & \textbf{161}    & 65.83    &  & 154                & 56.24              \\
PMS\_2021                  & 155                      &  & \textbf{100}    & 63.97    &  & 84                 & 51.65              \\
PMS\_2022                  & 179                      &  & \textbf{112}    & 69.07    &  & 97                 & 66.75              \\
WPMS\_2018                 & 172                      &  & \textbf{102}    & 105.23   &  & 82                 & 72.77              \\
WPMS\_2019                 & 297                      &  & \textbf{183}    & 95.78    &  & 155                & 93.41              \\
WPMS\_2020                 & 253                      &  & \textbf{157}    & 112.43   &  & 135                & 94.76              \\
WPMS\_2021                 & 151                      &  & \textbf{82}     & 136.22   &  & 61                 & 112.38             \\
WPMS\_2022                 & 197                      &  & \textbf{110}    & 121.01   &  & 84                 & 110.26      \\ \toprule      
\end{tabular}
\end{table}

\subsection{Ablation study}
We further perform ablation studies to analyze the effect of components and strategies in BandHS. The components include the \textit{hard} and \textit{soft} MAB models, and the HyDeci initialization method. The strategies mainly include the sampling strategy for selecting the candidate arms in the \textit{soft} MAB and the delayed reward method for updating the estimated values. All the studies show that our components and strategies in BandHS are effective and necessary.

\subsubsection{Ablation study on components}
To evaluate the effect of the \textit{hard } MAB, we perform two groups of experiments. The first group directly compares BandHS with BandMaxSAT~\cite{Zheng2022}, which only contains the \textit{soft} MAB (i.e., removes the \textit{hard } MAB from BandHS). The results are shown in Table \ref{table-BandHS-BandMS}. The second group compares BandHS$_{fast}$ with BandMaxSAT$_{fast}$, two variant algorithms of BandHS and BandMaxSAT, which output the first feasible solution they found (within a time limit of 300 seconds). The results are shown in Table \ref{table-BandHS-fast}, where column time$_1$ indicates the average running time in seconds to yield feasible solutions in solving each instance. For the instance that an algorithm can not yield feasible instances within 300 seconds, its running time is regarded as 300 seconds.

From the results in Tables \ref{table-BandHS-BandMS}, we can observe that BandHS outperforms BandMaxSAT on all the benchmarks, indicating that the proposed \textit{hard} MAB model can 
further boost 
the BandMaxSAT algorithm. From the results in Tables \ref{table-BandHS-fast}, we can observe that BandHS$_{\text{fast}}$ outperforms BandMaxSAT$_{\text{fast}}$ on most benchmarks and the average running time of BandHS$_{\text{fast}}$ to yield feasible solutions are shorter than that of BandMaxSAT$_{\text{fast}}$ on most benchmarks, indicating that the proposed \textit{hard} MAB model improves BandMaxSAT by finding better initial feasible solutions within a shorter time.

To evaluate the effect of the \textit{soft} MAB, we compare BandHS with its variant BandHS$_{\text{no-soft}}$, which simply sets the parameter $ArmNum$ to 1. This setting 
is equivalent to removing the \textit{soft} MAB from BandHS. The comparison results are shown in Table \ref{table-BandHS-soft}. We can observe that BandHS significantly outperforms BandHS$_{\text{no-soft}}$, indicating that the \textit{soft} MAB can help the local search algorithm find better feasible solutions.

Finally, we compare BandHS with its variant BandHS$_{\text{no-binary}}$ that does not prioritize binary clauses in HyDeci (i.e., remove lines 8-15 in Algorithm \ref{alg:HyDeci}). The results are summarized in Table \ref{table-BandHS-binary}. As the results show, BandHS shows much better performance than BandHS$_{\text{no-binary}}$ on WPMS instances, and shows lightly better performance than BandHS$_{\text{no-binary}}$ on PMS instances. This is because the local search in BandHS is more robust when solving PMS instances than solving WPMS instances, that can yield similar results with different initial solutions. The results also indicate that the proposed HyDeci algorithm that prioritizes both unit and binary clauses is effective and can improve the BandHS algorithm.

\subsubsection{Ablation study on strategies}
To evaluate the effect of the sampling strategy for selecting the candidate arms in the \textit{soft} MAB, we compare BandHS with its variant BandHS$_{\text{no-sample}}$, which selects the arm to be pulled in the \textit{soft} MAB by traversing all the available arms. The results are shown in Table \ref{table-BandHS-sample}, which demonstrates that the sampling strategy used in the \textit{soft} MAB model is effective and necessary.

We further compare BandHS with its variant BandHS$_{\text{no-delay}}$, which sets the parameter $d$ to 1, to evaluate the effect of the delayed reward method. The results are shown in Table \ref{table-BandHS-delay}. The results indicate that the delayed reward method fits well with the problems, and the method can help BandHS evaluate the quality of the arms better.






\section{Conclusion}
\label{Sec-Con}
In this paper, we propose a novel local search (W)PMS algorithm, called BandHS, which applies multi-armed bandit (MAB) models on hard and soft clauses respectively to guide the search directions when escaping from local optima. The \textit{hard} MAB is combined with all the literals in hard clauses to help BandHS select appropriate literals to satisfy the hard clauses, so as to find feasible solutions faster. The \textit{soft} MAB is combined with all the soft clauses to help BandHS select to satisfy appropriate soft clauses, so as to find better feasible solutions. We also propose an effective initialization method, called HyDeci, that prioritizes unit and binary clauses when constructing the initial solution. 

Extensive experiments show that our proposed methods have excellent performance and generalization capability, that significantly improve the state-of-the-art (W)PMS solvers, SATLike3.0 and NuWLS-c. Each component or strategy of BandHs, including the \textit{hard} MAB model, \textit{soft} MAB model, HyDeci initialization, sampling strategy in the \textit{soft} MAB, and the delayed reward calculation, is useful in improving the solution quality. In the future, we will further explore the potential of MAB methods in MaxSAT and SAT solving.



\section*{Acknowledgement}
This work is supported by National Natural Science Foundation of China (U22B2017) and Microsoft Research Asia (100338928).



\bibliographystyle{unsrt}

\bibliography{main}
\end{document}